\pdfoutput=1
\documentclass[10pt,final,conference,letterpaper,oneside,twocolumn]{IEEEtran}

\usepackage[cmex10]{amsmath}
\usepackage{graphicx}
\usepackage{url}
\usepackage{xspace}

\graphicspath{{images/}}
\DeclareGraphicsExtensions{.pdf,.jpg,.png}

\newcommand{\nop}{NimbRo\protect\nobreakdash-OP\xspace}
\newcommand{\dop}{DARwIn\protect\nobreakdash-OP\xspace}
\newcommand{\rviz}{RViz\xspace}
\newcommand{\nao}{Nao\xspace}
\newcommand{\nimbro}{NimbRo\xspace}
\newcommand{\scl}{State Controller Library\xspace}

\newcommand{\bcf}{Behavior Control Framework\xspace}

\newcommand{\secref}[1]{Section~\ref{sec:#1}\xspace}
\newcommand{\figref}[1]{Figure~\ref{fig:#1}\xspace}

\newcommand{\cpp}{C\texttt{\nolinebreak\hspace{-.05em}+\nolinebreak\hspace{-.05em}+}\xspace}

\usepackage{eso-pic}

\AtBeginDocument{\AddToShipoutPictureFG*{\AtTextUpperLeft{\put(0,\LenToUnit{21pt}){\parbox{\textwidth}{\centering\bfseries
Proceedings of 8th Workshop on Humanoid Soccer Robots, International Conference on\\Humanoid Robots (Humanoids), Atlanta, USA, 2013
}}}}}
\begin{document}

\bstctlcite{IEEEexample:BSTcontrol}

\title{A ROS-based Software Framework for the NimbRo-OP Humanoid Open Platform}

\author{\vspace{-3em} \\
\IEEEauthorblockN{Philipp Allgeuer, Max Schwarz, Julio Pastrana, Sebastian Schueller,
Marcell Missura and Sven Behnke\vspace{0.8em}}
\IEEEauthorblockA{Autonomous Intelligent Systems, Computer Science Institute VI, University
of Bonn, Germany\\
\tt\small \{pallgeuer, pastrana, missura\}@ais.uni-bonn.de\\\{schwarzm,
schuell1, behnke\}@cs.uni-bonn.de\\
http://ais.uni-bonn.de/nimbro/OP}} %

\maketitle

\begin{abstract}
Over the past few years, a number of successful humanoid platforms have been
developed, including the \nao~\cite{nao} and the \dop~\cite{dop}, both of which
are used by many research groups for the investigation of bipedal walking,
full-body motions, and human-robot interaction. The \nop is an open humanoid
platform under development by team \nimbro of the University of Bonn.
Significantly larger than the two aforementioned humanoids, this platform has
the potential to interact with a more human-scale environment. This paper
describes a software framework for the \nop that is based on the Robot Operating
System (ROS) middleware. The software provides functionality for hardware
abstraction, visual perception, and behavior generation, and has been used to
implement basic soccer skills. These were demonstrated at RoboCup 2013, as part
of the winning team of the Humanoid League competition.
\end{abstract}

\section{Introduction}
\label{sec:intro}

\begin{figure}
\centering
\includegraphics[width=0.95\linewidth]{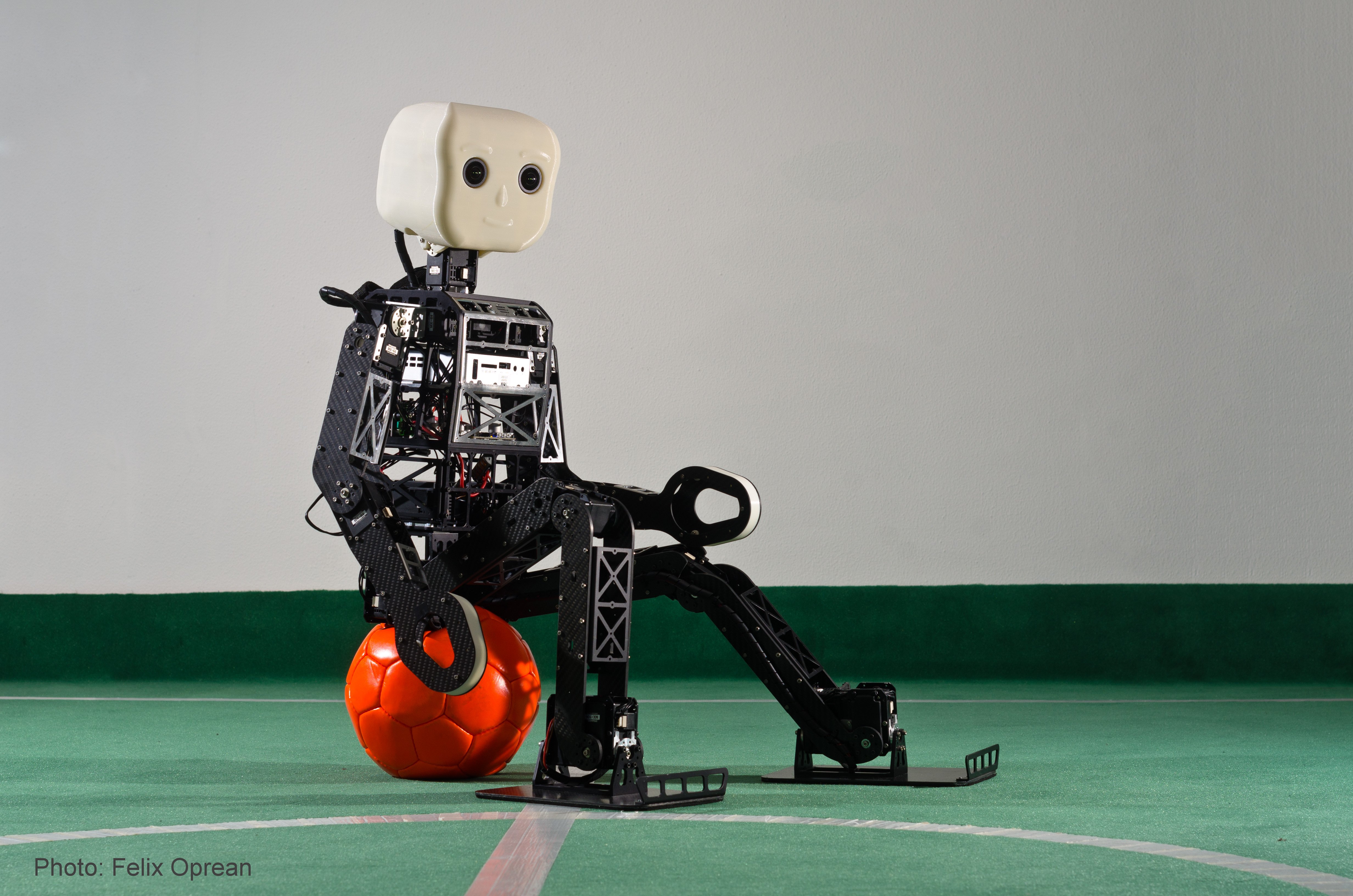}\vspace*{-1ex}
\caption{The \nop robot.}
\label{fig:noprobot}
\end{figure}

Progress in humanoid robotics research benefits from the
availability of robust and affordable platforms that provide a suitable
range of out of the box capabilities. Inspired by the success of the
\dop, a robot developed by the company Robotis, work commenced on the \nop in
early 2012. The first hardware prototype was displayed and demonstrated
at the IROS and Humanoids 2012 conferences. As an open platform, both the hardware
and software designs for the robot are committed to being regularly released
open source to other researchers and the general public. This hopes to foster
interest in the robot, and encourage research collaboration through
community-based improvement of the platform. Following the first hardware
release of the \nop, an initial software release was completed that was heavily
based on the framework released by Robotis for the \dop. This was however only a transitional
release pending an original and improved software framework based on the ROS
middleware. The second software release for the \nop, available at \cite{NOPSoftware},
constitutes exactly such a framework, and is referred to as the \nop ROS
Framework. The \nop robot is shown in \figref{noprobot}.

ROS \cite{ROSpaper}, an open source meta-operating system designed for robots,
provides features such as hardware abstraction, low-level device control,
packages for common functionalities, a communications framework, and a variety
of other libraries and tools. It is maintained by the Open Source Robotics Foundation, but is largely
a community project in terms of the contribution of packages. As the \nop robot
is currently predominantly being tested and demonstrated in the setting of
humanoid soccer (i.e.~RoboCup), numerous aspects of the framework are specific
to this domain. By virtue of the modular architecture of ROS and the
framework however, this does not prevent the robot from being used in other
settings. For example, the soccer nodes can simply be omitted from being
launched.

\section{Related Work}
\label{sec:relatedwork}

Due to its modularity, flexibility and portability, ROS has to date proven to be
a popular choice in the research community for the implementation of
robot control architectures. One of the main attractions of ROS is that it is so
widespread, and allows the work of different research groups from around the
world to be combined and shared in the form of ROS packages, even across robot
hardware barriers. This facilitates collaboration, and allows groups to both
benefit from the robotics research community, as well as contribute to it.
Examples of wheeled robots that have ROS integration include the PR2
and the TurtleBot. While ROS has many advantages due to its abstraction and modularity,
it also increases the communication requirements within the system, with data
such as vision detections and behavior commands needing to be transferred
via pipes.

Fewer examples of humanoid robots exist that have been integrated with ROS. For
example, the \dop nominally uses its own custom software framework, even though
this framework was written with future ROS integration in mind \cite{dop}. To
the knowledge of the authors, no open source ROS framework for
the \dop has become available to date. To a limited extent, a ROS framework for the HUBO
robot, a 130\,cm tall humanoid robot developed at the Korea
Advanced Institude for Science \& Technology, has been developed \cite{HUBOros}.
So far, this has only progressed to simple joint level control though, stopping
short of implementing ROS-compatible walking or manipulation tasks.

The widely-used \nao robot is a success story when it comes to ROS integration.
Based on an initial port by researchers at Brown University, the Humanoid Robots
Lab at the University of Freiburg has developed, and continues to maintain, a
collection of ROS packages and stacks that wrap the NaoQI API and make it
available in ROS. The UChile Robotics Team has published an extensive tutorial
on how to build, install and run ROS natively on the latest version of the \nao
\cite{naorostutorial}. This was believed to be the first initiative of its kind
for the \nao robot.

\section{Hardware Overview}
\label{sec:hardware}

\begin{figure}
\centering
\includegraphics[height=10cm]{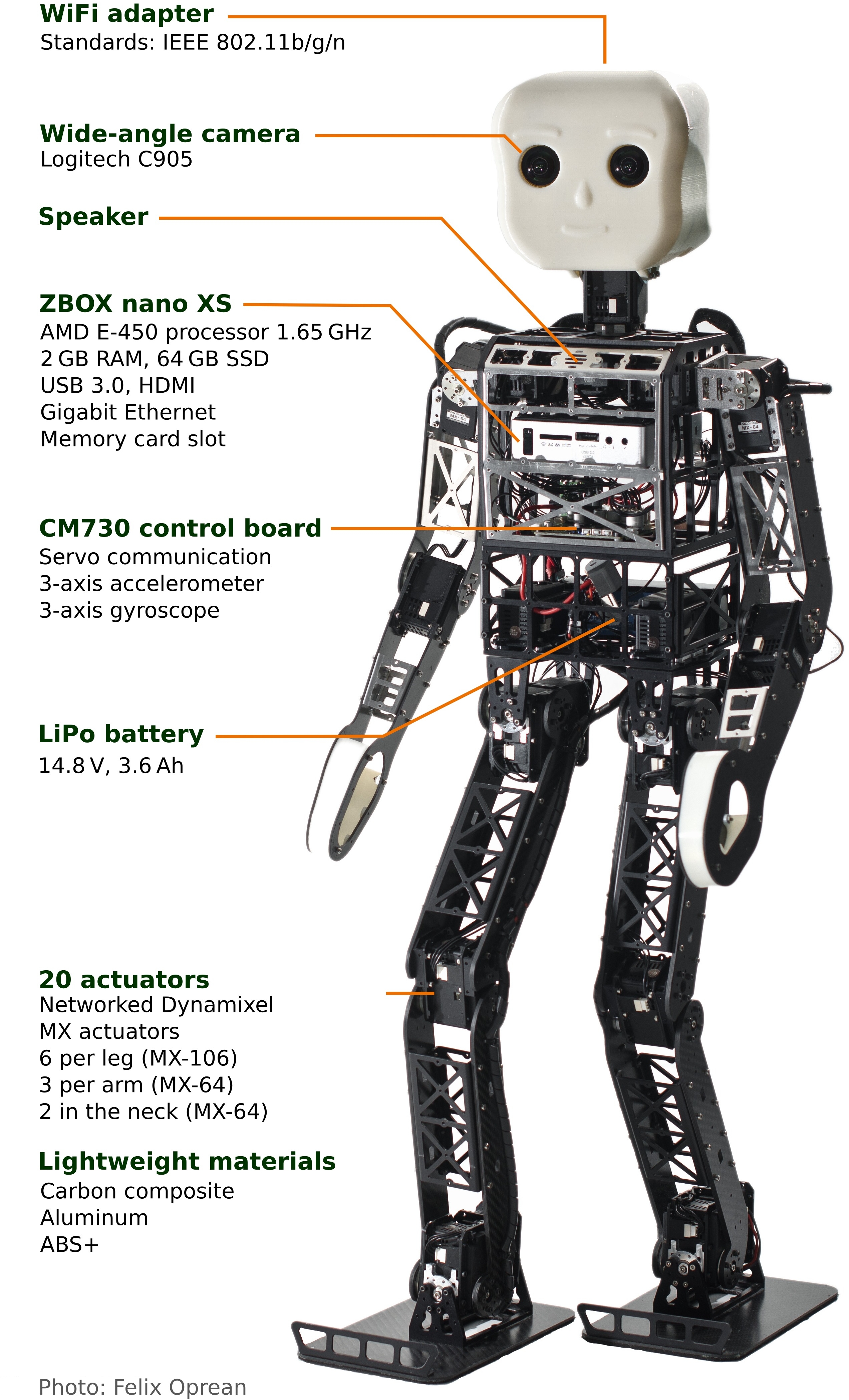} \hspace{6em}\vspace*{-1ex}
\caption{Diagram of the \nop robot hardware.}
\label{fig:nophardware}
\end{figure}

A diagram of the \nop that highlights all of the main hardware features is shown
in \figref{nophardware}. The robot is 95\,cm tall and weighs a total of 6.6\,kg
with the battery. This combination of height and
weight was chosen so as to allow the robot to be driven using only a single
actuator per joint. This decreases the
cost and complexity of the design, and allows the robot to exhibit quite large
ranges of motion in its 20 degrees of freedom. Robotis Dynamixel MX series
actuators are used for all of the joints, of which there are %
two in the neck, three in each arm and six in each leg. All of the Dynamixel
servos are chained together and addressed using a single one-wire TTL serial
bus.

At the heart of the robot is a Zotac Zbox nano XS PC, featuring a dual-core AMD
E-450 1.65\,GHz processor with 2\,GB RAM and a 64\,GB solid state disk, and
various communication interfaces including USB 3.0, HDMI and Gigabit Ethernet
ports. A USB WiFi adapter allows also for wireless networking. Data connections
to the robot automatically switch between TCP/IP and UDP depending on whether
the connection is wireless or not. The camera in the \nop's right eye, a
Logitech C905 model fitted with a wide-angle fisheye lens, is also connected to
the PC via USB. The extremely wide field of view is convenient for keeping
multiple objects in view at the same time.

A CM730 control board from Robotis, as used in their \dop robot, sits at the
approximate center of mass of the robot in the trunk. This board incorporates a
reprogrammable 32-bit ARM Cortex CPU and an inertial measurement unit, featuring
both a three-axis accelerometer and a three-axis gyroscope. Currently this CPU
is simply used to relay the sensory data to the computer and interface with the
servos. Three buttons and an array of multicolor LED lights are connected to the
CM730, and are available for custom use. As a non-standard addition, a
three-axis compass has been mounted in the trunk and interfaced to the CM730.
This allows the directional heading of the robot to be estimated. Appropriate
software for the anytime recalibration of the compass has been developed. More
detailed descriptions of the \nop hardware can be found in
\cite{NimbroOPDescription13} and \cite{NimbRoOPSWS}. 

\section{Software Architecture}
\label{sec:softwarearch}

\begin{figure*}
\centering
\vspace*{-1ex}
\includegraphics{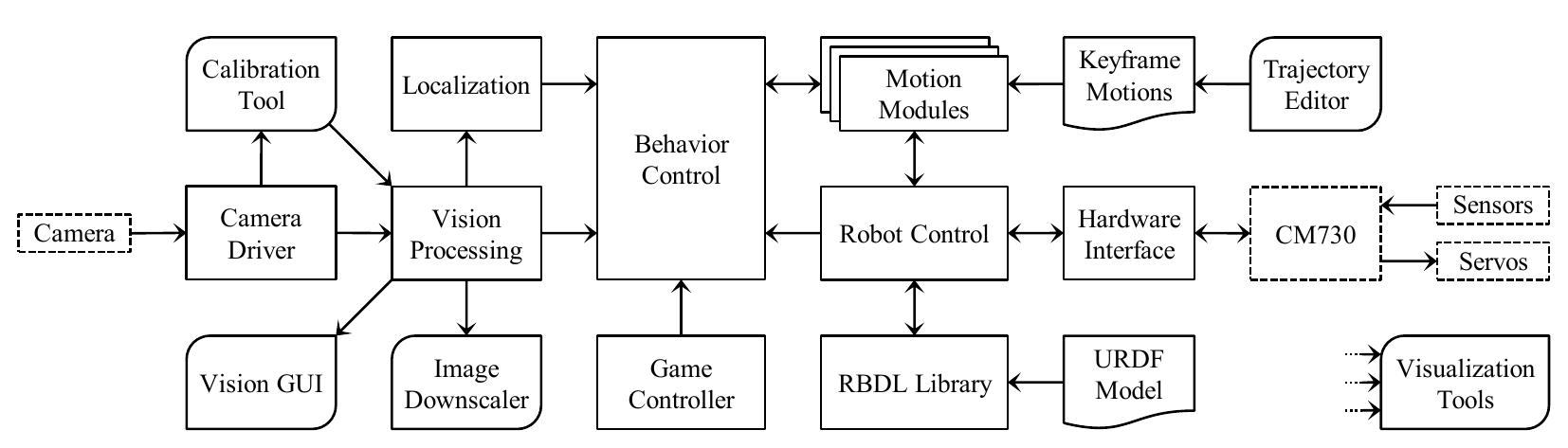}\vspace*{-.475ex}
\caption{Diagram of the \nop ROS framework software architecture. Boxes with a
dashed outline represent hardware elements, boxes with rounded corners represent
auxiliary software tools that are not necessarily required for execution of the
framework, and the arrows show the flow of information. Boxes with a wavy lower
edge represent data files. The visualization tools use information from
throughout the entire architecture.}
\label{fig:software_architecture}
\end{figure*}

The \nop software architecture makes use of ROS's native ability to split a
software solution into multiple independent processes called nodes. Each node
handles a certain aspect of the overall control task, and can be written,
compiled and launched separately from the other nodes if required. The nodes are
coupled using the ROS communication infrastructure and exchange information in
a direct peer-to-peer manner via ROS topics, service calls, shared memory, and
more. A decoupled architecture of this type is advantageous in that it can be
implemented in a very modular way, with relatively high independence between the
various elements of the system. Another advantage is that it naturally can be
split along the process boundaries and run on multiple cores, in a fashion
dynamically managed by the operating system scheduler.

The \nop software framework is implemented almost exclusively in \cpp, with only
miscellaneous use of additional scripting to perform tasks such as setting up a
computer for initial use of the framework, compiling the package documentation,
and providing framework-specific command line tools for everyday use. XML-based
launch scripts exist for all of the main tasks and configurations of the
framework, and are executed using the native ROS launch tool. The framework runs
on a Linux operating system, nominally Ubuntu. This system constraint
is predominantly imposed by the compatibility requirements of ROS itself.

A schematic of the \nop ROS framework software architecture is shown in
\figref{software_architecture}. The figure is generally arranged
such that the perception and state estimation tasks appear on the left, the
lower-level sensor and motion control tasks (i.e. Motion Control Layer; see
\secref{motionlayer}) appear on the right, and the higher level planning and
control tasks appear in the middle.

At the heart of the lower level control section is the \emph{Robot Control}
node, which contains the main real-time control loop that communicates with the
sensors and actuators, the camera excluded. This communication occurs via the
CM730 control board, and an exchangeable software component referred to as the
\emph{hardware interface}. The Robot Control node makes use of the RBDL library
\cite{RBDLLibrary} for various kinematic and dynamic calculations, which in turn
requires a model of the \nop in Unified Robot Description Format (URDF) format,
loaded from disk. Even though the servos are driven in the Robot Control node,
the coordination of the actual robot motions are left to the \emph{motion
modules}. One of these modules, a keyframe motion player, can execute motion
trajectories that are read from file. In control of the motion modules and
responsible for all higher level planning is the \emph{behavior control} node.
In the case of robot soccer, this reads game state data from the \emph{Game 
Controller} node.

A summary of the vision and localization nodes can be seen on the left-hand side
of \figref{software_architecture}. Images from the camera are made available via
the camera driver, and forwarded for processing by the color calibration tool
and vision processing node. Further tools include the vision graphical user
interface, and the image downscaler, described in \secref{vision} along with the
remainder of the vision system. The detections from the vision processing are
read by the behavior control node, and are used to drive the robot localization.

\section{Motion Control Layer}
\label{sec:motionlayer}

\subsection{Robot Control Node}
\label{sec:robotcontrol}

\begin{figure}
\centering
\vspace*{-2ex}
\includegraphics[width=0.95\linewidth]{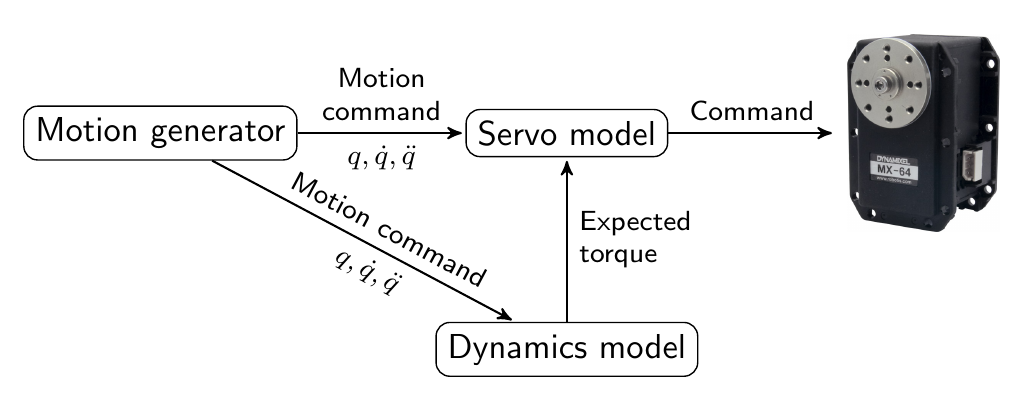}\vspace*{-1.5ex}
\caption{Schematic of the servo control architecture.}
\label{fig:servo_architecture}
\end{figure}

\begin{figure}
\centering
\includegraphics[width=0.95\linewidth]{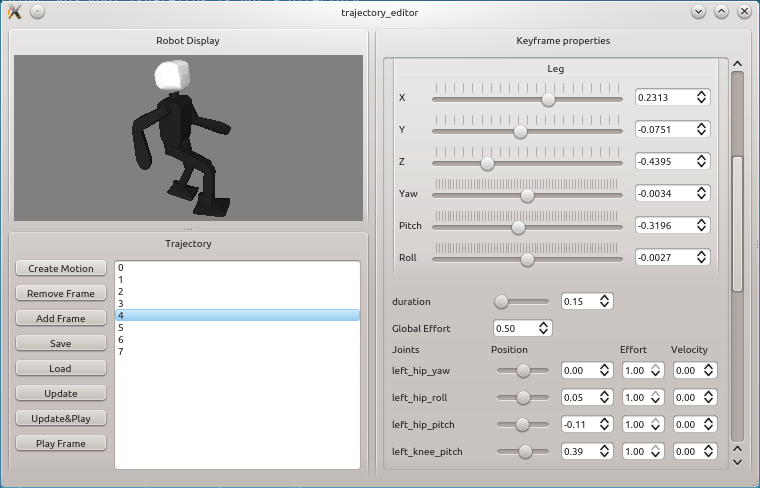}%
\caption{Screenshot of the trajectory editor, used to generate and edit the
keyframe motions used by the motion player.}
\label{fig:trajectory_editor}
\end{figure}

As indicated in \figref{software_architecture}, the Robot Control node is the
core of the Motion Control Layer. For the layers that come above it, it serves
to abstract away lower level details such as the internal robot model,
kinematics and dynamics. To preserve generality and portability, the Robot
Control node is extended via a plugin scheme to implement the hardware
interface. Two hardware interfaces have been developed within the scope of the
\nop project, one to implement the communication with the actual robot, and one
to simulate such communications with an adjustably noisy and time-delayed model.
The latter, referred to as the \emph{dummy interface}, is advantageous for
development in that code can be tested without risking damage to the robot.

Robot Control largely uses the RBDL library \cite{RBDLLibrary} for its kinematic
and dynamic calculations. In particular, the RBDL Newton-Euler inverse dynamics
algorithm is used to estimate the joint torques required for a particular
motion. These estimates are passed to a compliant servo controller, based on a
detailed model of the servos \cite{MaxRepcontrol}, that ensures that the robot
remains compliant while still following the target trajectories. This servo
control architecture is shown in \figref{servo_architecture}.

To avoid latencies, missed cycles and/or jitter, which can lead to
undesired robot behavior, the Robot Control loop was designed to be
as CPU and resource efficient as possible. Real-time performance of this node is
paramount, so it is given real-time privileges and uses the Linux high-precision
\emph{timerfd} API to ensure that timing constraints are met. The timer period
is nominally 8\,ms, giving rise to a nominal 125\,Hz update rate.

\subsection{Motion Modules}
\label{sec:motionmodules}

In addition to the plugin scheme implemented by Robot Control to
determine the robot hardware interface, a second plugin scheme was implemented
in the same node to import the actual motion code in a modular way. A plugin of
this second type is referred to as a \emph{motion module}. As opposed to being
hard-coded, the definition of which motion modules to load is provided at launch
time, allowing for greater flexibility and avoiding unnecessary recompilations.

The four main motion modules are the gait, head control, fall protection and
motion player modules. The \emph{gait module} implements an omnidirectional
walking gait for the \nop that is based on a central pattern generator. This
gait has been used by team \nimbro for numerous other humanoid robots \cite{Behnke2006}, and was
ported in this case for use in ROS and the \nop. The \emph{head control module}
is responsible for the motions of the neck servos, and ensures that the target
gaze directions commanded by the behavior control layer are effectuated. The
\emph{fall protection module} monitors the robot's state and triggers
the fall protection override when it detects that a fall is imminent, defined as
a threshold on the robot's attitude estimate. The override causes the servos to
relax and enter a free-wheeling mode, reducing the risk of damage to the gears.
Prone and supine get-up motions are used by the robot to recover once the fall
concludes~\cite{Stueckler_IAS_2006}. The final module, the \emph{motion player module}, is a keyframe
motion player that uses a nonlinear keyframe interpolation approach to achieve
smooth playback, and with great control. Keyframes are specified in terms of
both joint positions and velocities, and are chained together to form the
required motion(s) using timestamps. The interpolation scheme is based on
quadratic splines and takes velocity and acceleration limits into account. An
editor for these keyframe motions was created to facilitate their use, and is
referred to as the \emph{trajectory editor}. A screen shot of the trajectory
editor is shown in \figref{trajectory_editor}.

\section{Perceptions and State Estimation}
\label{sec:perceptions}

\subsection{Vision System}
\label{sec:vision}

\begin{figure}
\centering
\includegraphics[width=0.8\linewidth]{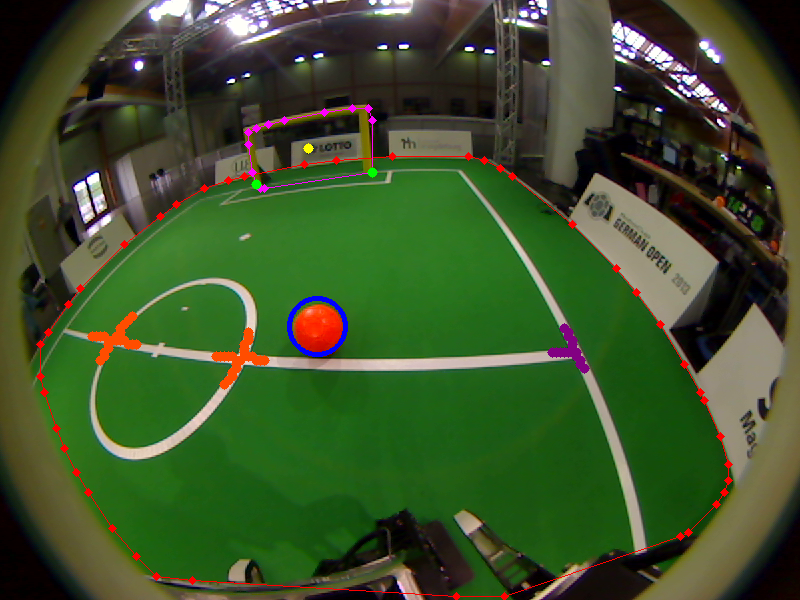}%
\caption{A snapshot from the \nop camera showing the detected features and the
distortion induced by the fisheye lens.}
\label{fig:cv_detections}
\end{figure}

\begin{figure}
\centering
\includegraphics[width=0.95\linewidth]{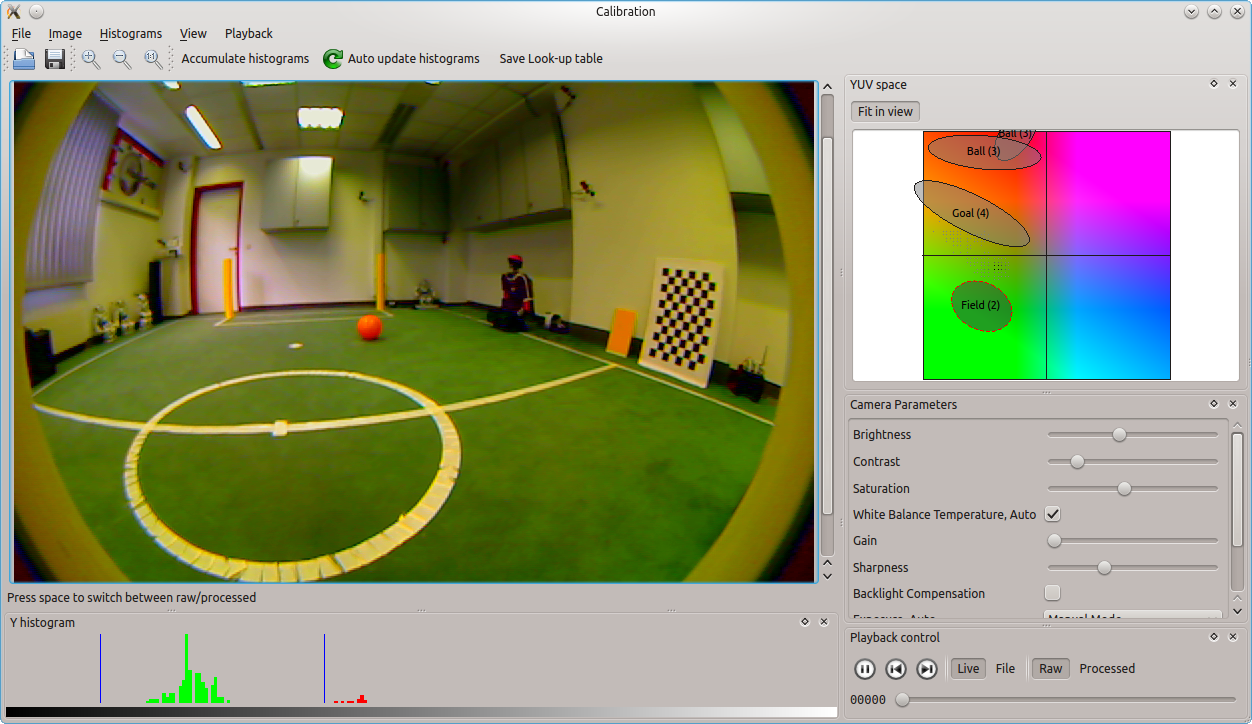}%
\caption{Color calibration tool used to generate YUV color lookup tables.}
\label{fig:cv_color_calib}
\end{figure}

The visual perception of a robot's surroundings is a crucial element in many
control systems, and in the case of the \nop, is a necessary prerequisite for
playing soccer. The \nop is equipped with a Logitech C905 camera, custom fitted
with a wide-angle fisheye lens to increase the field of view to approximately
180 degrees. This allows the robot to observe a large proportion of its
environment at any one time, and reduces the need for gaze control strategies to
locate objects. The Linux UVC camera driver is used to perform the image
acquisition via the \texttt{camera\_v4l2} plugin. Refer to
\figref{software_architecture} for a schematic of the vision software
architecture. The images are retrieved in YUYV format at a resolution of
$800\times600$, and at a rate of 24\,fps. Camera parameters such as brightness,
contrast and saturation can be adjusted using a custom color calibration tool,
shown in \figref{cv_color_calib}. This tool is also used to tune the color
lookup tables that are required for the subsequent color segmentation step (part
of the vision processing), which separates the image into orange, yellow, green,
white and black regions, and simultaneously subsamples the image by a factor of
four. The output of this step is a collection of subsampled binary images, where
each image corresponds to the pixel detections of a particular color. For the
purposes of robot soccer, these subsampled images are then used to detect the
extent of the field, and the required features of the environment such as
obstacles, line crossings, balls and goals. Finally, the pixel coordinates of
the object detections are undistorted into 3D Euclidean vectors that estimate
the location of the detected objects in egocentric world coordinates. A sample
image illustrating various object detections is shown in \figref{cv_detections}.

Two additional tools were written for the vision system. The first is a vision
GUI that can be used to inspect the raw, segmented and subsampled images, while
the second is an image downscaling node that is used to throttle and compress
images for publishing over WiFi. This allows images produced by the
running robot to be viewed on a remote computer.

\subsection{Localization}
\label{sec:localization}

The task of robot state estimation can be broken down into several components.
The pose of the robot---in terms of the configuration of all the
body parts relative to one another---is determined from the joint encoders via a
kinematic model of the robot and the corresponding forward kinematics. The
encoders are located directly on the joint axles, so they already incorporate
any servo non-linearities, in particular backlash, into their returned
measurements. The robot torso attitude is estimated based on 3D accelerometer
and gyroscope measurements by means of an angle estimation filter. This has been
demonstrated to provide a low noise, low latency and drift-free attitude
estimate. What remains is the requirement for an estimate of the global position
of the robot in its environment, in this case a soccer field. For this purpose, a
particle filter is used that processes the landmark detections from the vision
node, namely the detected white line crossings and goal posts, as well as the
compass readings. The implementation of localization on the \nop was ported from
the work in \cite{schulzlocalization}, and functions satisfactorily, but is
currently only in a beta stage. Future work includes the consideration of the
landmark orientations and the actual white line segments, as well as general
performance tuning of the particle filter.

\section{Configuration Server}
\label{sec:configserver}

Due to the large number of parameters in the \nop software, it is highly
unpractical or even infeasible to configure and tune the performance of the
robot if these parameters are incorporated in the source code as hard-coded
constants. Modifying hard-coded constants requires recompilation and
redeployment to the robot, potentially a very time-consuming process, and does
not allow for runtime changes to parameters. To address this issue, the
\emph{configuration server} was developed. Implemented as a ROS node, the
configuration server is a centralized manager and storage location for system
parameters. It is similar in idea to the native ROS parameter server, but with
many crucial improvements that make it suitable for use in the \nop software
framework. In particular, much effort was put into ensuring that the runtime
reconfigurability of the parameters is well supported, and that parameters can be
systematically shared between nodes. The existing \texttt{dynamic\_reconfigure}
ROS package, which extends the ROS parameter server, was insufficient for the
task as it did not allow the reconfigurable parameters to be shared globally between
the various nodes of the system. The parameters are stored in a hierarchical
structure in the server, with local copies of the relevant parameters being kept
in each of the nodes for performance reasons. The appropriate nodes are notified
via service calls whenever a parameter is changed on the server. Appropriate
functionality exists to be able to load and save the entire parameter hierarchy.
The default parameter values on system startup are taken from a well-defined
configuration file, stored on each robot.

\section{Behavior Control}
\label{sec:behavior}

\subsection{Behavior Frameworks}
\label{sec:behframeworks}

In order to perform tasks such as the playing of soccer, the \nop needs to be
able to be programmed with complex agent behaviors. Here, the term
\emph{behavior} is used to refer to an observable and coordinated pattern of
activity. As the behaviors constitute the higher-level control of the robot, it
is generally desired that such behaviors operate on a suitably abstracted view
of the environment, allowing the lower-level layers, such as the Motion Control
Layer described in \secref{motionlayer}, to take care of the specifics of the
walking and servo control aspects for example. Two generic cross-platform \cpp
frameworks were developed for the implementation of behaviors on the \nop
robot, the \scl and the \bcf \cite{behaviours}. The former is a state-based approach and
generalizes the concept of finite state machines to multi-action planning
derivatives thereof, while the latter is a behavior-based approach that
exploits a hierarchical structure and the concept of inhibitions to allow for
dynamic transitioning.

\subsection{Soccer Behaviors}

The soccer behaviors of the \nop are based on the \bcf, with the \scl being
used in a supporting role to implement finite state machines within the
individual behaviors if and as required. As the lower-level motions are already
handled by the Motion Control Layer, the soccer behaviors were split into only
three hierarchical \emph{behavior layers}, the soccer layer, the control layer,
and the ROS interface layer. Each non-interface layer contains a pool of robot
behaviors that compete dynamically via user-defined inhibitions for control of
the robot. The control layer contains skill-oriented behaviors such as
\emph{Dribble}, \emph{Kick} and \emph{Go Behind Ball}, the execution of which
are resolved at runtime based on the game state and the perceptions of the
robot. The soccer layer on the other hand, contains behaviors of even greater
abstraction, including in particular \emph{Play Soccer}. The purpose of the
soccer layer is to accommodate the inclusion of additional behaviors, that can
address any auxiliary aspects of the soccer game, such as the technical
challenges at RoboCup. The ROS interface layer is a special behavior layer with
no child behaviors. Instead, it handles the communication of the input and
output data of the behavior node over the ROS topics and services. As this
allows the implementation of the remaining two behavior layers to be
ROS-independent, this makes the behaviors more portable to other systems.

In terms of behaviors, the \nop is currently capable of searching for the ball,
walking behind it while taking into account any perceived obstacles, dribbling
it towards the target half of the field, and kicking the ball towards the goal.
Additional behaviors of the control layer that execute in parallel to this
include a game control behavior that processes the signals of the RoboCup game
controller during a match, and a head control behavior, which ensures that the
robot's camera is always pointing in a suitable direction during the game. It is
to be noted however, that despite all that has been achieved in terms of
behavior control so far, there will be more advanced behaviors to come in the
future software releases.

\section{System Visualization}
\label{sec:visualization}

\begin{figure}
\centering
\includegraphics[width=1.0\linewidth]{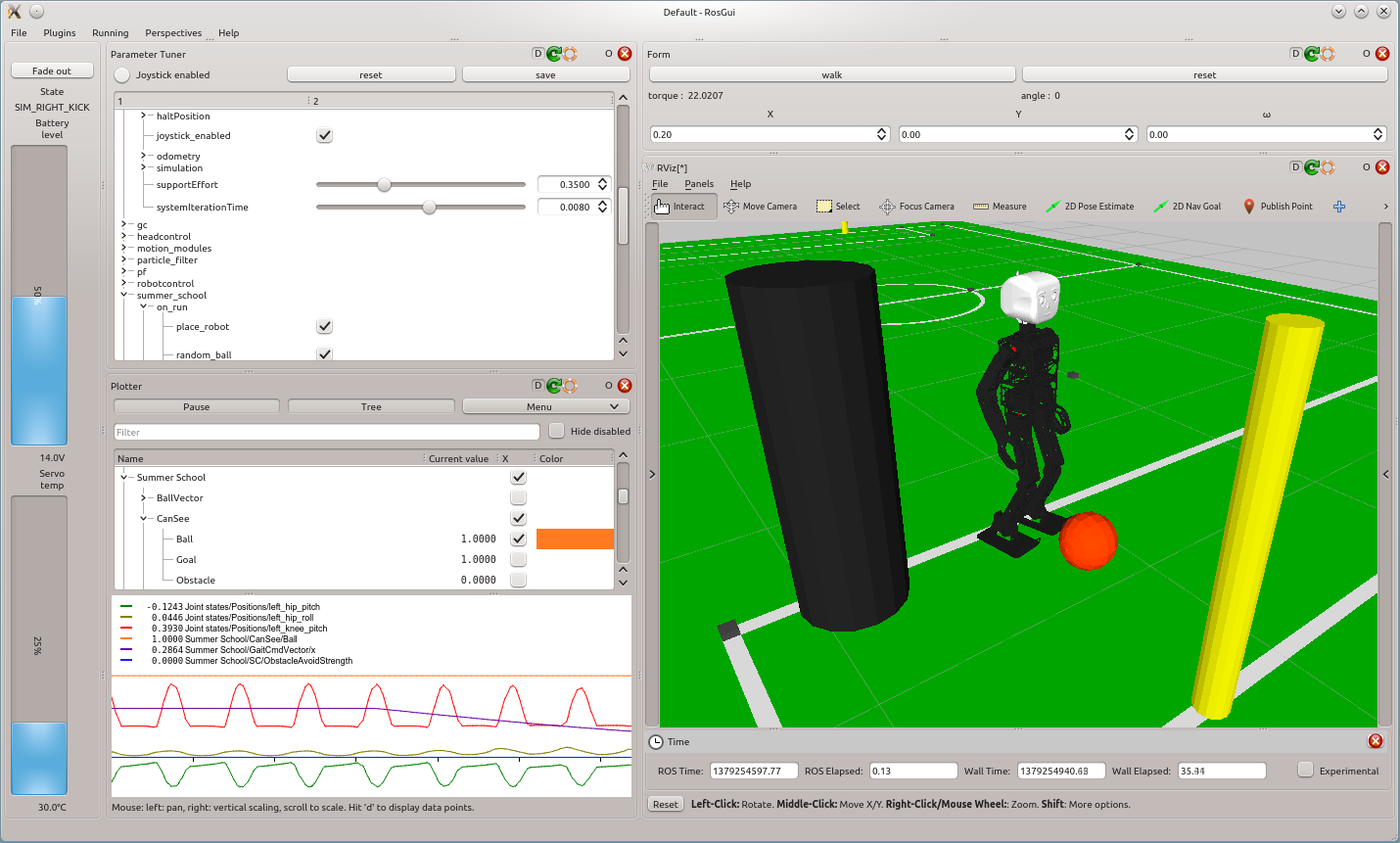}%
\caption{Screenshot of the RQT visualization tool, showing the \rviz, walk
control, plotter, parameter tuner and diagnostics widgets.}
\label{fig:visualizationsnap}
\end{figure}

Visualizations are indispensable tools at all stages of the code development
cycle, in particular for the purposes of debugging and testing. They can also
relay valuable state information to the operator during normal runtime
situations. The visualization tools for the \nop are all implemented as widgets
for the ROS-standard RQT graphical user interface (GUI), and can be seen in
\figref{visualizationsnap}. The most fundamental of these widgets is the \rviz
plugin, which provides a highly customizable 3D visualization environment within
which the robot can be displayed.
The robot is shown in either its current or desired pose, as specified by its
joint angles, and at the global location it currently thinks it is at. A virtual
soccer field environment can be enabled to
augment the view. Many additional features can be displayed in the 3D virtual
environment, including custom point markers, line traces, footstep locations,
servo torques, link coordinate frames, and object detections such as ball, goal,
obstacle and landmark detections. These allow the user to get a quick visual
insight into the state of the robot, and detect any obvious problems or
discrepancies. 

The diagnostics widget, the leftmost widget in \figref{visualizationsnap},
displays the robot's current battery voltage level and maximum servo
temperature. It also displays the current motion state, as defined in Robot
Control and the motion modules, and allows the user to fade the servos in and
out. The walk control widget, located above \rviz in the
figure, allows the user to issue manual gait commands and otherwise control the
gait module, provided that the currently executing behavior code is not writing
values of its own. This is useful for instance, for the testing of the Motion
Control Layer. The parameter tuner widget allows the user to view, organize,
modify, load and save parameters of the configuration server. Numeric parameters
are represented and can be modified using sliders and spin boxes, while
checkboxes are used for boolean parameters. This permits the user to manually
access the configuration server and modify aspects of the system at runtime,
either for testing or tuning purposes.

The final visualization that can be seen in \figref{visualizationsnap} is the
plotter widget. This is a plugin that can be used to trace and plot the time
history of any nominated variables throughout the entire framework code. The
node in which the variable appears just needs to publish its value to the
plotter, where it is automatically placed into a circular buffer and made
available for display to the user. The plotter variables are stored in a tree
hierarchy, which is presented to the user so that a selection of variables to
plot can be made. An important feature of the plotter is that the mouse cursor
can be used to view the robot's state at a past time. This affects what is shown
in the 3D visualization. The RQT tool had to be modified in order to allow the negative
time changes that are observed when the user scrolls back in time. This
extension to RQT is encapsulated in the so-called \emph{timewarp} package.
Finally, at any point in time the plotted data can be flushed to a ROS bag using
the save feature of the widget.

\section{Conclusion}
\label{chap:conclusion}

In this paper a ROS-based software framework was presented for the \nop robot, a
humanoid open platform in development by team \nimbro of the University of
Bonn. This is the second software release to be undertaken for this robot, but
the first to feature the ROS middleware. All dependencies and inclusions of the
original Robotis software framework have been rewritten and/or removed. The
decision to release the framework open source was so as to support collaboration
and community-based improvements.

The \nop ROS framework has been demonstrated publicly on numerous occasions,
including at ICRA 2013 and RoboCup 2013, where it scored its first official
soccer competition goal. The framework was also demonstrated in an earlier form
at the RoboCup German Open 2013, where the \nop received the HARTING Open-Source
Award. Work on the \nop is still an ongoing project however, so accordingly the
ROS framework presented in this paper is expected to remain subject to further
revisions and additions. Future work on the framework includes the
incorporation of a multi-hypothesis Kalman filter for object filtering and
tracking, the integration of a physics-based simulator for testing, and the development of more
advanced soccer behaviors. Additional goals include the development of a more
advanced gait, and the tuning of the localization for improved robustness.

\section*{Acknowledgement}

This work was partially funded by grant BE 2556/10 of the German Research
Foundation (DFG).

\bibliographystyle{IEEEtran}
\bibliography{ms}

\end{document}